\documentclass[dvipsnames]{article}
\usepackage{paperbenchx}

\usepackage[utf8]{inputenc}
\usepackage[T1]{fontenc}
\usepackage{amsmath}
\usepackage{booktabs,tabularx,longtable}
\usepackage{graphicx}
\usepackage{float}
\usepackage{caption}
\usepackage{enumitem}
\usepackage{listings}
\usepackage[most]{tcolorbox}
\usepackage{pifont}
\usepackage{xspace}
\usepackage{xurl}
\usepackage{siunitx}
\usepackage{dblfloatfix}
\usepackage{placeins}

\newcommand{\benchmarkname}{\textsc{PaperBenchX}\xspace}
\newcommand{\bloglink}{https://unipat.ai/blog/PaperBenchX}
\newcommand{\leaderboardlink}{https://unipat.ai/benchmarks/PaperBenchX}
\newcommand{\nAreas}{12}
\newcommand{\nSimulators}{10}
\newcommand{\nTasksCur}{93}
\newcommand{\nResultTasks}{93}
\newcommand{\nRubricTasks}{93}
\newcommand{\nLeaves}{3{,}168}
\newcommand{\nModels}{10}
\newcommand{\nExpertHours}{1{,}000}
\newcommand{\code}[1]{\begingroup\urlstyle{tt}\path{#1}\endgroup}
\newcommand{\cmark}{\textcolor{ForestGreen}{\ding{51}}}
\newcommand{\xmark}{\textcolor{BrickRed}{\ding{55}}}
\definecolor{taskbackground}{RGB}{247,248,250}
\definecolor{taskframe}{RGB}{75,85,99}
\lstdefinestyle{paperbenchxcode}{
  basicstyle=\ttfamily\footnotesize,
  breaklines=true,
  columns=fullflexible,
  keepspaces=true,
  showstringspaces=false,
  frame=single,
  rulecolor=\color{taskframe},
  backgroundcolor=\color{taskbackground}
}

\title{\benchmarkname: Can AI Reproduce Science Across Disciplines?}

\author{
  {\fontsize{10pt}{11pt}\selectfont UniPat AI}
}

\renewcommand{\partnerlogo}{
  \raisebox{0.9ex}{\color[HTML]{4D4D4D}\scshape\small
  \DTMdisplaydate{2026}{10}{29}{-1}}}

\begin{document}

\maketitle

\begin{abstract}

\textbf{Before AI can be trusted to discover new science, it should be able to reproduce what science already knows.} Reproduction provides a rigorous checkpoint for an AI scientist: the target claim is known, yet recovering it still requires translating an incomplete scientific description into an executable experiment and producing evidence that can be independently verified. Existing research-reproduction benchmarks largely focus on AI and machine learning, where experiments share relatively homogeneous software stacks. Extending this capability across scientific disciplines introduces a different challenge: agents must operate domain-specific scientific software, make physical and numerical choices, diagnose long-running computations, and distinguish valid evidence from merely plausible outputs.
We introduce \textbf{\benchmarkname}, a benchmark of 93 expert-constructed paper-reproduction tasks spanning 12 research areas across electromagnetics, photonics, chemistry, materials science, biology, and robotics. Each task provides the source paper, a public task specification, and a domain-native scientific environment involving specialized scientific software. Agents must reconstruct a key experiment, execute and diagnose it, and deliver a runnable workflow that supports the target scientific claim.
To evaluate scientific validity rather than surface-level agreement, we replay every submission from scratch and score only regenerated evidence against expert-validated criteria. Across 10 frontier agent configurations, the best configuration fully reproduces only 14\% tasks under the 90\% threshold. Meanwhile, mean rubric scores range from 27.53\% to 59.08\%, indicating that substantial partial progress rarely translates into complete scientific reproduction. These results reveal a large gap between paper understanding and reliably producing reproducible scientific evidence.
 \end{abstract}

\begingroup
\setlength{\intextsep}{6pt}
\setlength{\abovecaptionskip}{4pt}
\setlength{\belowcaptionskip}{0pt}
\begin{figure}[H]
  \centering
  \includegraphics[width=\textwidth]{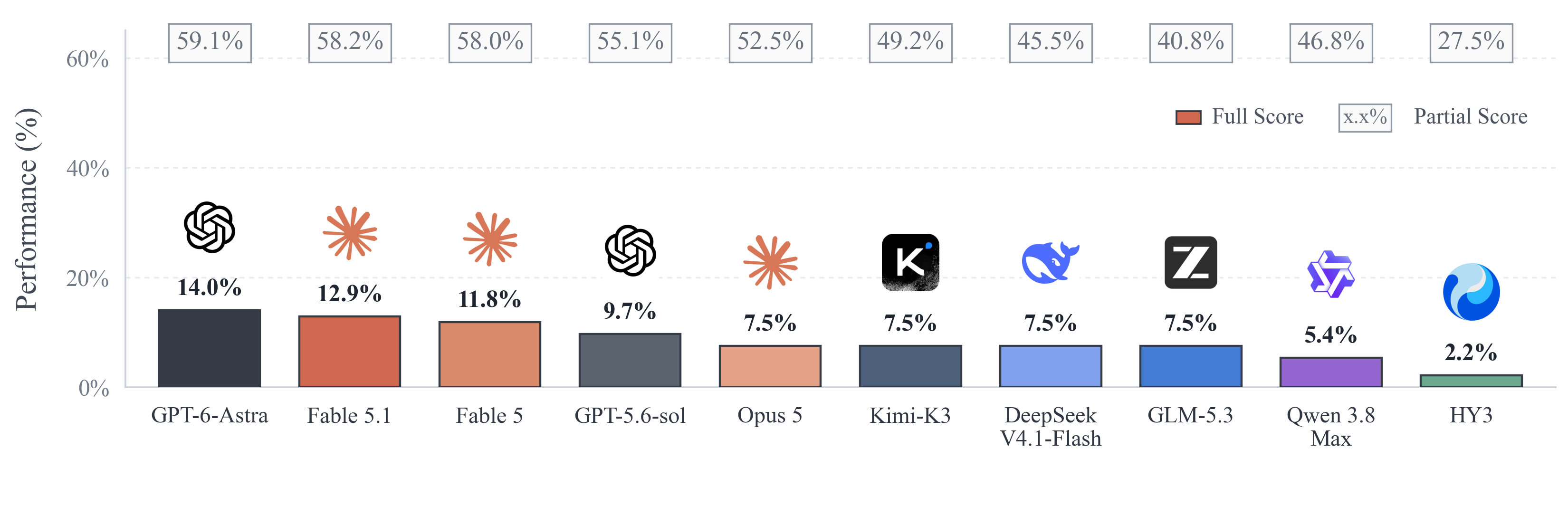}
  \caption{\textbf{Frontier agents on \benchmarkname.} Bars show full scores
  ($>90\%$); boxes show partial scores.
  GPT-6-Astra and GPT-5.6-sol use Codex; all others use Claude Code.}
  \label{fig:leaderboard}
\end{figure}
\endgroup

\section{Introduction}
\label{sec:introduction}

Reproducibility is one of the defining features of science: a result is meaningful not only because it has been reported, but because the assumptions, procedures, and evidence behind it can be reconstructed and independently examined. For AI systems, this makes reproduction a demanding test of scientific competence. An agent must go beyond understanding a paper's conclusion to recover how the result was obtained, translate incomplete descriptions into executable procedures, operate specialized tools, diagnose failures, and determine whether the resulting evidence actually supports the claim. Existing reproduction benchmarks largely focus on AI and machine learning, while cross-domain scientific benchmarks typically isolate only parts of the workflow~\citep{starace2025paperbench,siegel2024corebench,geng2026scienceide}. \textbf{What remains underexplored is whether an agent can reproduce published experiments across disciplines, inside the environments native to the underlying science.}

This distinction matters because, outside machine learning, the environment is not merely infrastructure. It is part of the scientific problem. Reproducing an antenna result may require constructing and solving an electromagnetic model in HFSS; reproducing a photonics result may depend on geometry, boundary conditions, mode extraction, and convergence behavior in Lumerical or Meep~\citep{oskooi2010meep}; reproducing a chemistry or materials result may require choosing basis sets, numerical tolerances, or electronic-structure settings; and reproducing a robotics result may require executing and validating behavior inside an embodied simulator. A workflow can run successfully, generate plausible numbers, and still represent the wrong physical system. Cross-disciplinary scientific reproduction therefore demands more than writing code that executes.

\begin{figure*}[t]
  \centering
  \includegraphics[width=\textwidth]{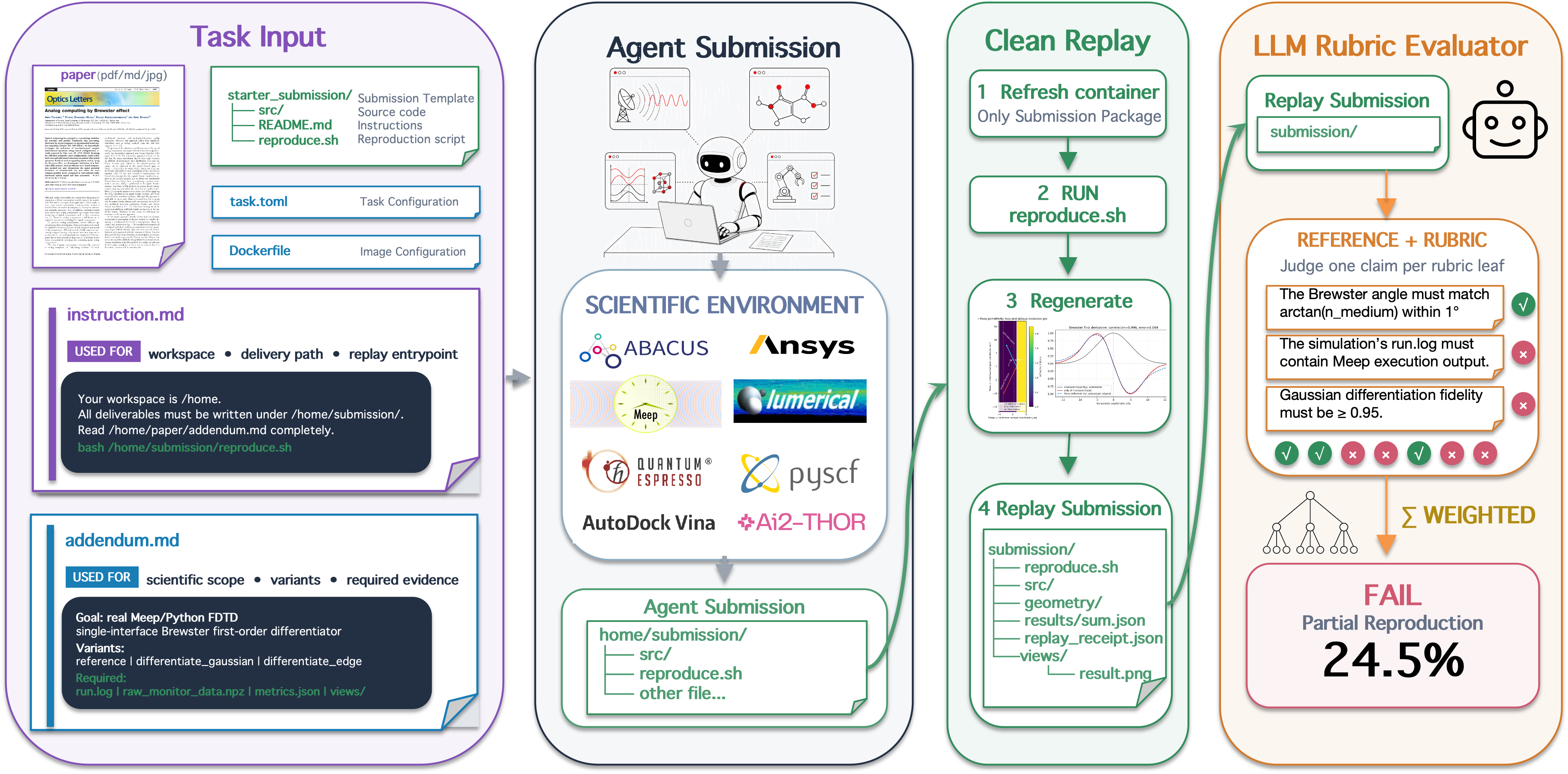}
  \caption{\textbf{From task package to replay-grounded score.}
  An agent receives the paper, task specification, environment, and starter
  submission; interacts with domain-native scientific software; and returns a
  runnable package. Clean replay regenerates the evidence used to score the
  scientific rubric. The displayed leaves and scores are illustrative.}
  \label{fig:task-replay}
\end{figure*}

We identify three coupled challenges. \textbf{First, scientific reconstruction:} papers are not executable specifications. Agents must translate prose, equations, figures, and domain conventions into a concrete experiment, identify the target observable, and resolve modeling choices that are not fully specified by the task contract. \textbf{Second, domain-native execution:} agents must operate specialized scientific software, select appropriate physical and numerical settings, interpret diagnostics, and determine whether a computation is stable and converged. \textbf{Third, scientific verification:} matching a reported value is not sufficient. The result must arise from a valid experimental workflow, and the evidence supporting it must be independently regenerable. These challenges separate scientific reproduction from either rerunning released code or producing a plausible final answer.

To study this setting, we introduce \textbf{\benchmarkname{}}, a benchmark for paper-level scientific reproduction across disciplines. It contains 93 expert-constructed tasks spanning 12 research areas across electromagnetics, photonics, computational chemistry, computational materials science, computational biology and bioinformatics, and robotics and embodied AI. The tasks run in 10 domain-native scientific computing environments, including Ansys HFSS, Ansys Lumerical, Meep~\citep{oskooi2010meep}, PySCF/GPU4PySCF~\citep{sun2026pyscf}, ABACUS~\citep{zhou2025abacus}, Quantum ESPRESSO~\citep{giannozzi2009qe}, AutoDock Vina~\citep{eberhardt2021vina}, Habitat~\citep{savva2019habitat}, SimplerEnv~\citep{li2025simpler}, and AI2-THOR~\citep{kolve2017ai2thor}. Each task is centered on a verifiable scientific claim from a published paper. The agent receives the paper, a public task specification defining the reproduction scope and deliverables, and the required scientific environment, but does not receive the private reference workflow, reference evidence, or evaluation rubric. Its goal is to reconstruct the experiment and deliver a runnable workflow that regenerates evidence supporting the target claim.

\textbf{\benchmarkname{}} evaluates sustained, end-to-end scientific reproduction through replay-grounded verification: agents must maintain scientific intent, computational state, and evidence provenance while reconstructing experiments, running and diagnosing long-running simulations, and refining their analyses; after submission, each workflow is rerun from scratch in a clean environment, and only regenerated artifacts are scored against expert-validated criteria. This design distinguishes plausible implementations and successful software execution from actual reproduction of the target scientific result. Across 10 frontier agent configurations on the same 93 tasks, the strongest configuration exceeds the 90\% full-reproduction threshold on only 13 tasks, while 71 tasks are not fully reproduced by any evaluated configuration; mean rubric scores range from 27.53\% to 59.08\%. Agents often make substantial progress in modeling and execution but fail at scientific validation, revealing a substantial gap between partial progress and reliably closing the loop from paper understanding to reproducible scientific evidence.

Our contributions are threefold:
\begin{itemize}[leftmargin=1.5em,nosep]

\item \textbf{Cross-disciplinary paper-level reproduction.} We introduce 93 expert-constructed tasks spanning 12 research areas and 10 scientific computing environments, each centered on a verifiable claim from a published paper.

\item \textbf{Domain-native scientific execution.} Rather than reducing scientific problems to a common coding interface, \benchmarkname{} requires agents to operate inside the simulators, software stacks, numerical conventions, and resource constraints required by the underlying research.

\item \textbf{Replay-grounded scientific verification.} We rerun every submitted workflow from scratch and score only regenerated evidence against expert-validated scientific criteria, distinguishing plausible outputs from reproducible scientific results.

\end{itemize}

\section{Related Work}
\label{sec:related-work}

\begin{table*}[!b]
  \centering
  \small
  \caption{\textbf{Comparison with prior agent benchmarks.} Checkmarks indicate
  that a benchmark explicitly evaluates the named capability.}
  \label{tab:related-benchmarks}
  \setlength{\tabcolsep}{4pt}
  \renewcommand{\arraystretch}{1.04}
  \resizebox{\textwidth}{!}{%
  \begin{tabular}{lllccc}
    \toprule
    \textbf{Benchmark}
      & \textbf{Research scope}
      & \textbf{Primary evaluation target}
      & \shortstack{\textbf{Paper-level}\\\textbf{reproduction}}
      & \shortstack{\textbf{Domain-native}\\\textbf{scientific execution}}
      & \shortstack{\textbf{Replay-grounded}\\\textbf{scientific verification}} \\
    \midrule
    \multicolumn{6}{l}{
      \textit{Machine-learning research and paper-reproduction benchmarks}
    } \\
    \addlinespace[2pt]
    MLAgentBench~\citep{huang2023mlagentbench}
      & Machine learning
      & ML experimentation tasks
      & \xmark & \xmark & \xmark \\
    RE-Bench~\citep{wijk2024rebench}
      & Machine learning
      & Research-engineering tasks
      & \xmark & \xmark & \xmark \\
    PaperBench~\citep{starace2025paperbench}
      & Machine learning
      & Paper reproduction
      & \cmark & \xmark & \xmark \\
    SA-Bench~\citep{hu2026sabench}
      & Machine learning
      & Paper reproduction
      & \cmark & \xmark & \cmark  \\
    \midrule
    \multicolumn{6}{l}{\textit{Cross-domain scientific benchmarks}} \\
    \addlinespace[2pt]
    DiscoveryBench~\citep{majumder2024discoverybench}
      & 6 scientific fields
      & Data-driven discovery
      & \xmark & \xmark & \xmark \\
    ScienceAgentBench~\citep{chen2024scienceagentbench}
      & 4 scientific fields
      & Scientific workflow tasks
      & \xmark & \xmark & \xmark \\
    CORE-Bench~\citep{siegel2024corebench}
      & 3 scientific fields
      & Repository-result reproduction
      & \xmark & \xmark & \xmark \\
    DISCOVERYWORLD~\citep{jansen2024discoveryworld}
      & 8 scientific fields
      & Interactive scientific discovery
      & \xmark & \cmark & \xmark \\
    NatureBench ~\citep{wang2026naturebench}
      & 6 scientific fields
      & Scientific experiment tasks
      & \xmark & \xmark &  \cmark  \\
    ScienceIDE~\citep{geng2026scienceide}
      & Physics and geoscience
      & Repository-level scientific coding
      & \xmark & \cmark & \cmark  \\
    \midrule
    \textbf{\benchmarkname}
      & \textbf{\nAreas{} research areas}
      & \textbf{Reproduction of key paper experiments}
      & \cmark & \cmark & \cmark \\
    \bottomrule
  \end{tabular}}
\end{table*}

\subsection{Machine-Learning Research and Paper Reproduction}

MLAgentBench and RE-Bench evaluate long-horizon machine-learning
experimentation and research engineering, but their objectives are specified as
task-level metric improvement rather than reproduction of published
claims~\citep{huang2023mlagentbench,wijk2024rebench}. PaperBench and SA-Bench
move toward paper-level evaluation through reproduction or semantic-alignment
rubrics~\citep{starace2025paperbench,hu2026sabench}. They establish that a final
metric alone cannot measure fidelity to a paper, but remain centered on
relatively homogeneous machine-learning software stacks rather than
heterogeneous domain-native scientific environments.

\subsection{Cross-Domain Scientific Benchmarks}

Cross-domain benchmarks broaden the scientific scope but generally isolate only
part of the reproduction process. SciCode, DiscoveryBench, and
ScienceAgentBench evaluate scientific coding, hypothesis discovery, or data
analysis, while CORE-Bench recovers results by rerunning released
code~\citep{tian2024scicode,majumder2024discoverybench,
chen2024scienceagentbench,siegel2024corebench}. DISCOVERYWORLD evaluates
interactive discovery in a unified virtual world, and NatureBench constructs
paper-derived scientific-ML tasks~\citep{jansen2024discoveryworld,
wang2026naturebench}. ScienceIDE most directly evaluates domain-native
scientific execution by placing agents inside production scientific software,
but its tasks concern repository-level repair and implementation rather than
reproduction of a published experiment~\citep{geng2026scienceide}.

\benchmarkname{} targets the intersection left open by these benchmarks. It
requires \textbf{paper-level reproduction of key experiments}, executes tasks
inside heterogeneous \textbf{domain-native scientific environments}, and uses
\textbf{replay-grounded scientific verification}. Submitted workflows must run
again from scratch and regenerate evidence that supports the target claim;
neither a plausible final answer nor a completed repository is sufficient.
Table~\ref{tab:related-benchmarks} summarizes these distinctions.

\section{The \benchmarkname{} Benchmark}
\label{sec:paperbenchx}

\benchmarkname{} evaluates whether an agent can turn a scientific claim into
independently replayable evidence in the domain-native environment required by
the underlying research. Figure~\ref{fig:task-replay} summarizes the complete
lifecycle, from the materials provided to the agent through clean replay and
evidence-based scoring. This section defines the task, describes the benchmark
composition and construction pipeline, and specifies the evaluation protocol.

\begin{figure*}[!b]
  \centering
  \includegraphics[width=\textwidth]{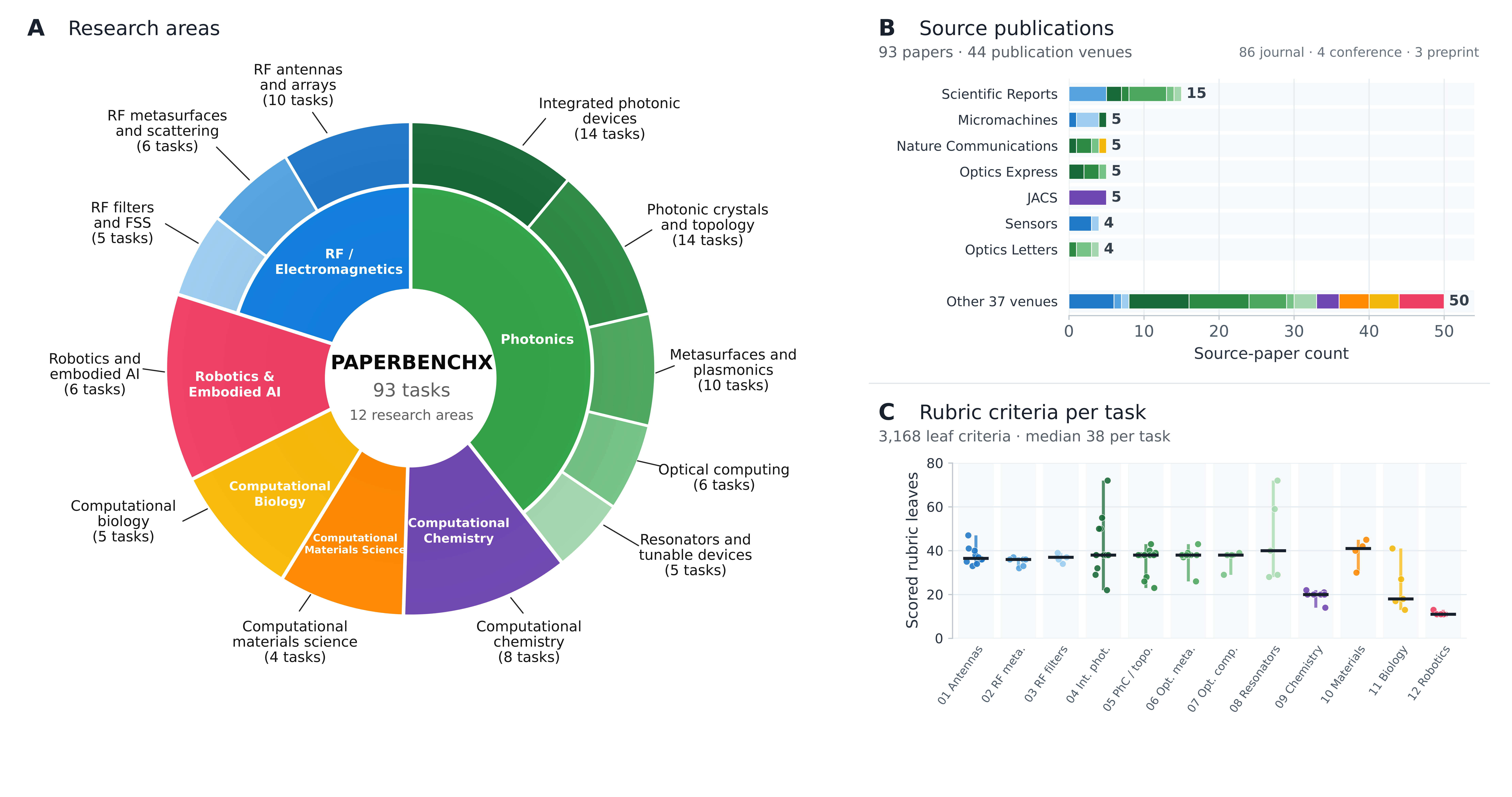}
  \caption{\textbf{Coverage and evaluation structure of \benchmarkname.}
  \textbf{A}, The 93 tasks span 12 research areas in six scientific domains.
  \textbf{B}, Their 93 source papers span 44 publication venues; bar colors
  correspond to the 12 research areas in A. \textbf{C}, The task rubrics contain
  3,168 scorable criteria (median 38 per task); points are tasks and black ticks
  mark research-area medians.}
  \label{fig:task-area-distribution}
\end{figure*}

\subsection{Task Definition}
\label{sec:design}

Each \benchmarkname{} task is centered on a verifiable scientific claim from a
published scientific paper. As shown in Figure~\ref{fig:task-replay}, the task
input has three parts: the paper; a public task specification defining the
scientific scope, target conditions and variants, observables, required
evidence, and delivery interface; and a scientific computing environment
containing the domain-native simulator, dependencies, data, and allocated
resources. The environment is part of the task definition rather than
incidental infrastructure, because model representations, numerical
conventions, execution procedures, and convergence evidence differ across
scientific software.

The agent must submit a runnable package containing all source, model-building,
simulation, post-processing, and analysis code needed for reproduction, with
\code{reproduce.sh} as the canonical entry point. During clean replay, the
evaluator refreshes the task container, restores only the submitted package,
runs this entry point, and collects the outputs regenerated by the workflow.
Those outputs, rather than the agent's assertions, are scored against a hidden
scientific rubric. The private reference workflow and \emph{reference
artifacts} are never exposed to the agent. Reference artifacts are
expert-generated simulator logs, raw numerical outputs, convergence records,
and derived metrics produced during task construction; they establish that the
task is executable and ground its expected evidence and scoring criteria.
Values reported in the source paper remain visible, but private tolerances and
reference evidence are withheld. The task therefore tests whether an agent can
construct a scientifically valid and independently replayable procedure, not
whether it can recover a hidden number.

\subsection{Task Statistics}
\label{sec:statistics}

\begin{figure*}[!t]
  \centering
  \includegraphics[width=\textwidth]{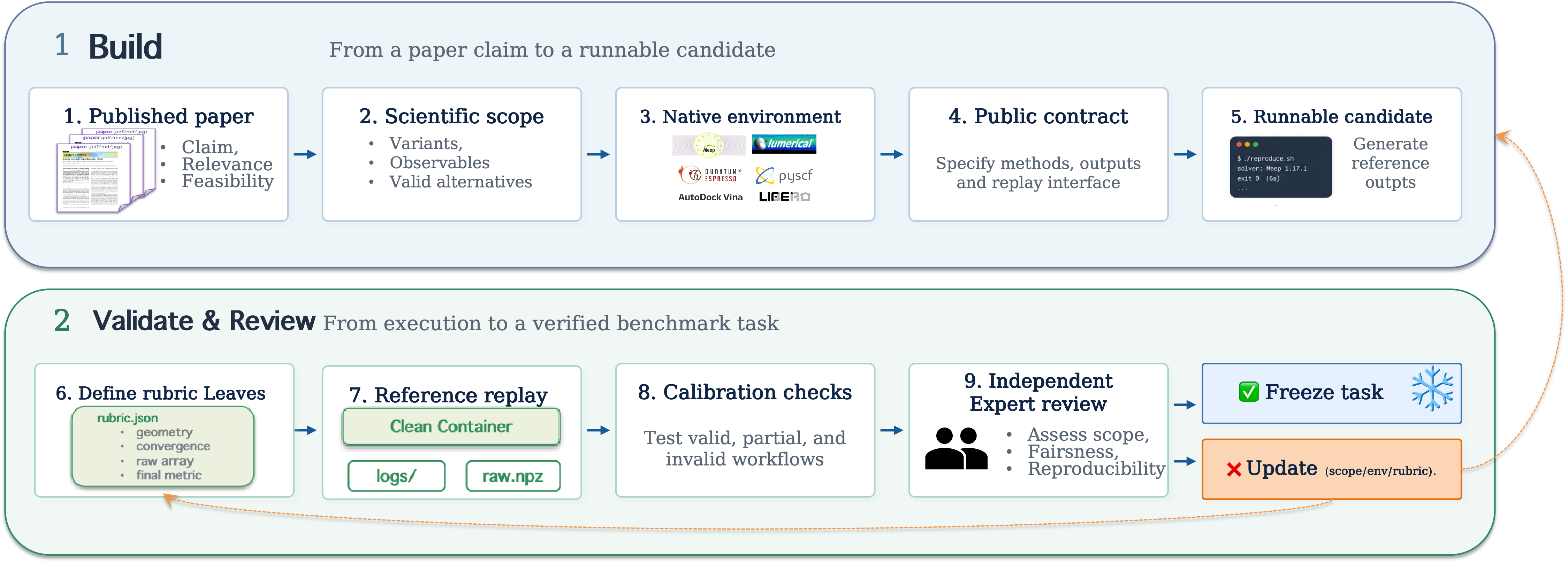}
  \caption{\textbf{Task construction and quality control.}
  Experts turn a published claim into a runnable task by defining its scientific
  scope, provisioning the domain-native environment, and specifying the public
  task contract. Reference replay, calibration checks, and independent expert
  review verify executability, fairness, and evidence requirements before the
  task is admitted to the benchmark.}
  \label{fig:task-construction}
\end{figure*}

\paragraph{Benchmark composition.}
\benchmarkname{} contains \nTasksCur{} paper-reproduction tasks spanning
\nAreas{} research areas and \nSimulators{} domain-native scientific computing
environments, with \nLeaves{} scorable criteria. The twelve research areas are
radio-frequency antennas and arrays; radio-frequency metasurfaces and
scattering; radio-frequency filters and frequency-selective surfaces;
integrated photonic devices; photonic crystals and topological photonics;
optical metasurfaces, gratings, and plasmonics; optical computing and signal
processing; optical resonators, filters, and tunable devices; computational
chemistry; computational materials science; computational biology and
bioinformatics; and robotics and embodied AI.
Figure~\ref{fig:task-area-distribution}A shows the distribution of tasks across
these research areas.

The tasks run in ten domain-native scientific computing environments: Ansys
HFSS, Ansys Lumerical, Meep, PySCF/GPU4PySCF, ABACUS, Quantum ESPRESSO,
AutoDock Vina, Habitat, SimplerEnv, and AI2-THOR. The benchmark draws on 93
distinct source papers from 44 publication venues, as summarized in
Figure~\ref{fig:task-area-distribution}B. Each task has an execution budget
determined by its simulation scale and resource requirements; agent time limits
range from \SI{4}{\hour} to \SI{72}{\hour}, with a median of \SI{7}{\hour}.
Figure~\ref{fig:task-area-distribution}C shows the distribution of scorable
criteria across the same \nRubricTasks{} tasks.

\paragraph{Release and benchmark integrity.}
We release one representative task from each of the twelve research areas.
Nine tasks include complete, redistributable scientific computing environments;
the remaining three rely on commercially licensed software, for which we
release the task specifications and scoring details but not the corresponding
software environments. The remaining tasks are retained as a hidden evaluation
set to reduce test-set contamination and preserve the benchmark's validity for
evaluating future agents.

\subsection{Task Construction Pipeline}
\label{sec:tasks}

The pipeline proceeds in two stages (Figure~\ref{fig:task-construction}): first
building a runnable candidate task, and then validating whether it can be
admitted to the benchmark.

\noindent\textbf{Stage 1: Build.} \textbf{Steps 1--2: paper and claim
selection.} A domain expert selects a
published paper, identifies a scientifically meaningful claim, and checks that
it can be reproduced under the available software, licensing, compute, and time
constraints. The expert then defines the scientific scope: the target figure or
table, model configuration, experimental conditions and variants, observables,
valid methodological alternatives, and evidence required to support the claim.
\textbf{Steps 3--4: environment and public contract.} The expert provisions the
domain-native scientific environment and writes the public task specification,
which states the scope, required outputs, submission location, and replay entry
point without disclosing the private rubric or reference solution.
\textbf{Step 5: initial execution check.} The expert implements a private
reference workflow and executes it in the assembled environment. This first
execution check verifies that the simulator can be invoked, the scoped
experiment can finish within the resource envelope, and the required reference
artifacts can be generated. A task reaches the candidate stage only after this
end-to-end workflow runs successfully.

\noindent\textbf{Stage 2: Validate and review.} \textbf{Step 6: rubric
construction.} The expert converts the evidence
requirements into independently scorable rubric leaves covering model
construction, parameter settings, execution, convergence, post-processing, and
claim validation. Each leaf specifies the evidence needed for credit and its
scoring rule, tolerance, and weight. \textbf{Step 7: clean reference replay.}
The private reference workflow is rerun from a clean copy of the final
container through the same replay interface used for submissions. Unlike the
initial check in Step 5, this stricter execution check verifies that the task
does not depend on stale state, undeclared files, or hidden setup; it also
confirms that the expected artifacts can be regenerated under the final
resource limits and consumed by the evaluator. \textbf{Step 8: calibration
checks.} Successful, partially successful, and invalid workflows are scored to
confirm that the rubric awards supported progress, withholds credit for missing
or contradictory evidence, and distinguishes a complete reproduction from a
plausible but scientifically invalid result. \textbf{Step 9: independent expert
review.} Two independent domain experts inspect the scientific scope, fairness,
reproducibility, evidence requirements, and scoring behavior.

Tasks that fail because of their scope, environment, or reference workflow are
rebuilt from Stage 1. Failures caused by the rubric, scoring rules, or evaluator
return to Step 6. In either case, all affected downstream execution,
calibration, and review checks are repeated. Only tasks that pass the full
pipeline are frozen and admitted to the benchmark. Across the benchmark, task
construction and validation required more than \nExpertHours{} hours of
doctoral-level expert work, or approximately ten hours per task.

\vspace{-1.2ex}
\subsection{Evaluation}
\label{sec:verification}
\vspace{-0.7ex}

Scientific correctness cannot be established from an agent's final report.
\benchmarkname{} therefore evaluates each submission through clean replay,
evidence-based grading, and hierarchical scoring. This evaluation of the
agent's workflow is distinct from the clean \emph{reference} replay in Step 7
of task construction, which validates the benchmark task itself.
Appendix~\ref{app:protocol} gives the full protocol.

\noindent\textbf{Clean replay.} After delivery, generated outputs and mutable
execution state are removed. The
evaluator refreshes the task container, restores only the agent's submitted
package, and invokes \code{bash /home/submission/reproduce.sh} with private
evaluation resources inaccessible, outbound network disabled, and
task-specific resource limits enforced. Exit status, timing, source hashes,
and execution provenance are recorded independently of the agent's report.
Only artifacts regenerated by this replay are admissible evidence; a replay
failure is recorded explicitly, and credit remains possible only for rubric
requirements supported by admissible regenerated evidence.

\noindent\textbf{Evidence-based grading.} Replay outputs---including simulator
logs, raw numerical arrays, trajectories,
convergence records, derived metrics, and visualizations---are frozen as a
read-only evidence set. Every scientific conclusion must be traceable to this
set: a convergence claim requires a convergence record, and a reported
observable requires both the raw output and the analysis that derives it.
Deterministic checks evaluate directly testable properties such as file schema,
units, numerical consistency, and task-specific tolerances. Requirements that
need domain interpretation are assessed for every evaluated configuration by
the same GPT-5.6-sol scientific rubric evaluator, with access to the hidden
rubric, reference artifacts, and read-only replay evidence.
The agent's report may locate or explain evidence, but it cannot instruct the
evaluator or replace missing or contradictory artifacts.

\noindent\textbf{Rubric stages.} Each rubric leaf is assigned to
\emph{Modeling} (scientific models, assumptions, inputs, parameters, and
configuration), \emph{Execution} (domain-native execution, required completion
or convergence, artifacts, and provenance), or \emph{Validation}
(post-processing and quantitative support for the target claim). Stage scores
normalize earned weight by the total rubric weight within each stage.

\noindent\textbf{Scoring.} Let $s_i \in [0,1]$ be the score of rubric leaf $i$
and $w_i$ its effective hierarchical weight. The leaves represent independently
assessable scientific requirements, and their effective weights sum to one.
The task score is
\vspace{-0.6ex}
\begin{equation}
  R = \sum_i w_i s_i, \qquad \sum_i w_i = 1.
  \label{eq:task-score}
\end{equation}
\vspace{-0.8ex}
We report performance at both the task-completion level and the partial-progress
level using two metrics:
\begin{itemize}[leftmargin=1.2em,nosep]
  \item \emph{Full rate}, the fraction of tasks with $R>0.9$. A score of exactly
  $0.9$ does not qualify.
  \item \emph{Partial score}, the mean of $R$ across the shared \nResultTasks{}-task result set,
  which captures weighted progress on incomplete long-horizon reproductions.
\end{itemize}
Every configuration is evaluated on the same task set; experiment-specific
fallback and missing-outcome handling is defined in
Section~\ref{sec:experimental-setup}. A missing artifact affects only the rubric
leaves that depend on it, leaving unrelated verified evidence eligible for credit.

\section{Experiments}
\label{sec:experiments}

\subsection{Experimental Setup}
\label{sec:experimental-setup}

\paragraph{Models and scaffolds.}
We evaluate ten model--scaffold configurations:
GPT-6-Astra (Codex)~\citep{openai2026models},
Fable~5.1~\citep{anthropic2026models},
Fable~5~\citep{anthropic2026models},
Opus~5~\citep{anthropic2026models},
GPT-5.6-sol (Codex)~\citep{openai2026models},
Kimi-K3~\citep{moonshot2026kimik3},
Qwen3.8-Max~\citep{alibabacloud2026modelcatalog},
GLM-5.3~\citep{zhipu2026glm53,alibabacloud2026modelcatalog},
DeepSeek-V4.1-Flash~\citep{deepseek2026v41flash}, and
HY3~\citep{tencent2026hunyuan}. Tasks are packaged as Harbor
environments~\citep{harbor2026}. The two GPT models use
Codex~\citep{openai2026codex}, while the other eight use Claude
Code~\citep{anthropic2026claudecode}. All configurations use the maximum
available reasoning setting and receive the same public task materials and
shell access, without access to the private reference workflow or rubric.

\paragraph{Evaluation protocol.}
All configurations are evaluated on the same \nResultTasks{} tasks. When
Fable~5 or Fable~5.1 refuses a task, the system automatically falls back to
Opus~5: the original refusal is retained for the refusal rate, while the
fallback submission determines the task score. Missing outcomes and refusals
without a valid fallback count as zero. We report the \emph{partial score} as
the mean rubric score, the \emph{full rate} as the fraction scoring strictly
above 90\%, and the \emph{refusal rate} from the original model responses. All
available submissions are clean-replayed and graded by the same GPT-5.6-sol
evaluator~\citep{openai2026models}, supplemented by deterministic checks.

\subsection{Main Results}
\label{sec:main-results}

\begin{table}[!htbp]
  \centering
  \setlength{\tabcolsep}{3pt}
  \renewcommand{\arraystretch}{1.03}
  \resizebox{\linewidth}{!}{%
  \begin{tabular}{@{}llrrrrrrrr@{}}
    \toprule
    & & \multicolumn{3}{c}{Overall (\%)}
      & \multicolumn{3}{c}{Stage score (\%)}
      & \multicolumn{2}{c}{Average effort} \\
    \cmidrule(lr){3-5}\cmidrule(lr){6-8}\cmidrule(l){9-10}
    Model & Harness & Partial & Full & Ref.
      & Modeling & Execution & Validation & Turns & Time (h) \\
    \midrule
    GPT-6-Astra          & Codex       & \textbf{59.08} & \textbf{13.98} & 0.00  & \textbf{72.72} & 69.35 & \textbf{52.76} & 348.7 & 5.96 \\
    Fable 5.1            & Claude Code & 58.17 & 12.90 & 11.83 & 71.00 & 69.92 & 51.06 & 113.0 & 4.65 \\
    Fable 5              & Claude Code & 57.96 & 11.83 & 13.98 & 72.13 & 68.99 & 50.39 & 112.0 & 5.09 \\
    GPT-5.6-sol          & Codex       & 55.07 & 9.68  & 0.00  & 69.85 & \textbf{71.23} & 45.29 & 237.8 & 3.28 \\
    Opus 5               & Claude Code & 52.52 & 7.53  & 0.00  & 66.99 & 67.16 & 47.76 & 259.2 & 6.52 \\
    Kimi-K3              & Claude Code & 49.24 & 7.53  & 0.00  & 62.40 & 60.70 & 42.32 & 185.9 & 5.75 \\
    Qwen3.8-Max          & Claude Code & 46.76 & 5.38  & 0.00  & 64.41 & 61.02 & 41.65 & 234.4 & 5.76 \\
    DeepSeek-V4.1-Flash  & Claude Code & 45.47 & 7.53  & 0.00  & 59.87 & 57.51 & 41.13 & 372.6 & 6.74 \\
    GLM-5.3              & Claude Code & 40.78 & 7.53  & 0.00  & 51.11 & 51.85 & 37.06 & 290.8 & 6.84 \\
    HY3                  & Claude Code & 27.53 & 2.15  & 0.00  & 36.83 & 40.59 & 19.67 & 197.4 & 4.94 \\
    \midrule
    Average              & --          & 49.26 & 8.60 & 2.58 & 62.73 & 61.83 & 42.91 & 235.2 & 5.55 \\
    \bottomrule
  \end{tabular}%
  }
  \caption{\textbf{Model-level results on the shared set of 93 tasks.}
  Partial, Full, Ref., and all stage scores are reported as percentages;
  metric definitions and outcome handling follow
  Section~\ref{sec:experimental-setup}.}
  \label{tab:model-results}
\end{table}

\paragraph{Overall performance.}
GPT-6-Astra (Codex) leads with a $59.08\%$ partial score and a $13.98\%$
full rate, yet exceeds the $90\%$ threshold on only 13 of 93 tasks. Across
models, the macro-average partial score is $49.26\%$, whereas the average full
rate is only $8.60\%$, showing that substantial partial progress rarely becomes
complete reproduction. This gap is concentrated in scientific validation:
average Modeling and Execution scores reach $62.73\%$ and $61.83\%$,
respectively, while Validation falls to $42.91\%$. Current agents can therefore
make substantial progress in modeling and execution, but reliably reproducing
a paper's key experimental results and producing a scientific workflow
sufficient to verify them remains difficult.

\begin{figure}[!htbp]
  \centering
  \includegraphics[width=0.95\linewidth]{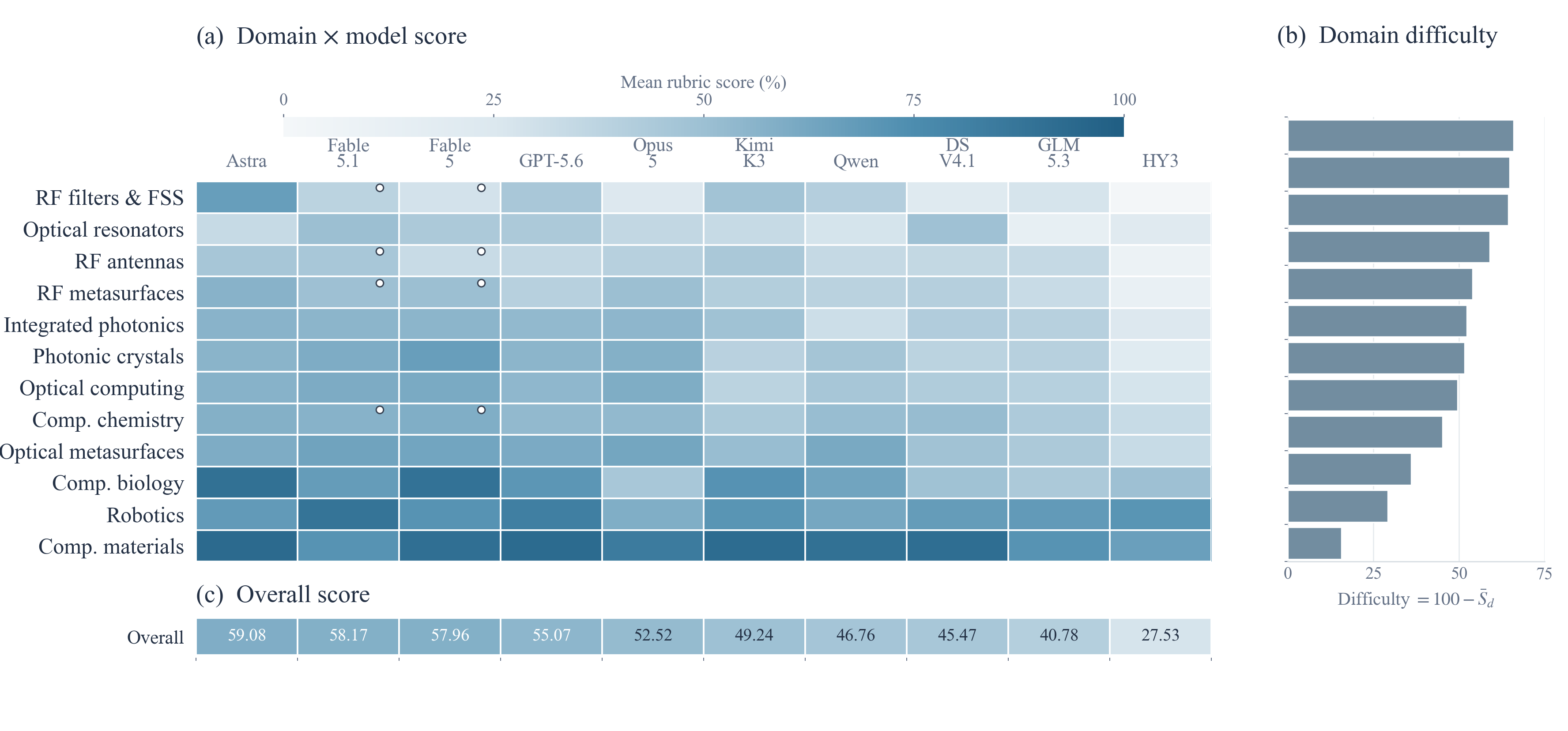}
  \caption{\textbf{Domain-specific capability profiles.}
  \textbf{(a)} Mean rubric score by model and research area, ordered from
  hardest to easiest; circles mark Fable refusals scored using the Opus~5
  fallback. \textbf{(b)} Domain difficulty, defined as 100 minus the
  cross-model mean. \textbf{(c)} Overall scores, matching
  Table~\ref{tab:model-results}. Names are abbreviated.}
  \label{fig:area-matrix}
\end{figure}

\paragraph{Performance by Research Area.}
Performance varies substantially across research areas
(Figure~\ref{fig:area-matrix}). Mean partial score ranges from $34.05\%$ for
RF filters and frequency-selective surfaces to $84.32\%$ for computational
materials science. The
heatmap also reveals substantial model--domain variation: GPT-6-Astra leads six
of the twelve areas, while Fable~5 and Fable~5.1 each lead three, and no
configuration dominates every area.

\section{Analysis}
\label{sec:discussion}

\begin{figure*}[t]
  \centering
  \includegraphics[width=\textwidth]{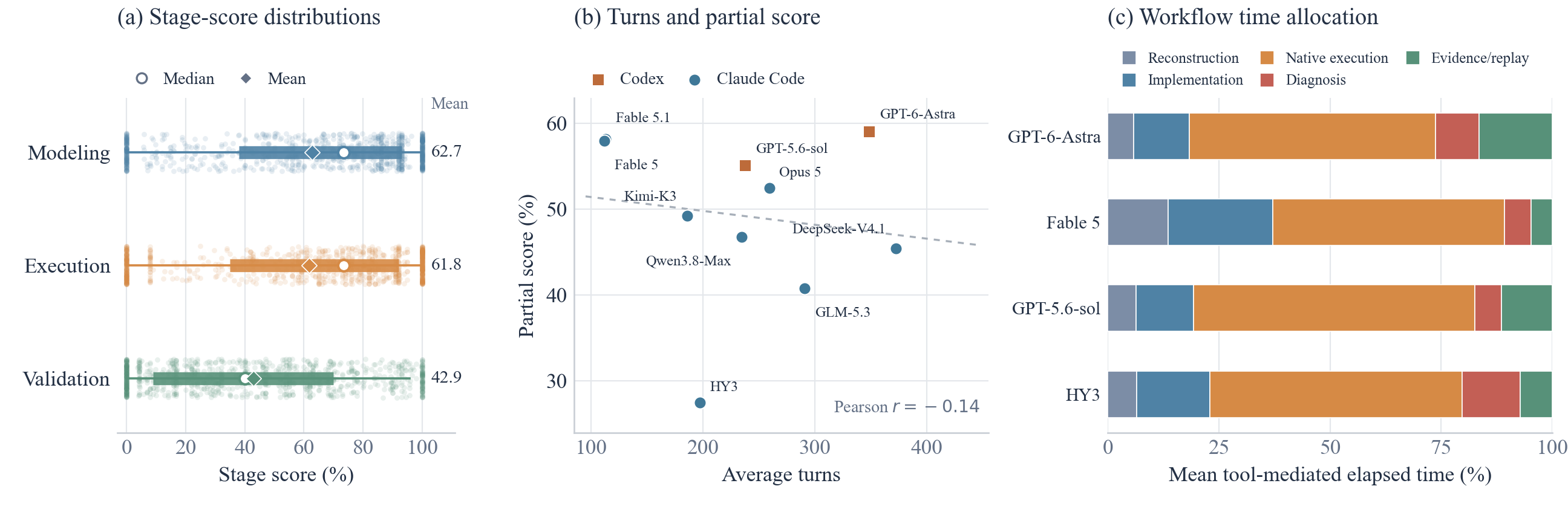}
  \caption{\textbf{Stage performance, interaction length, and workflow allocation.}
  \textbf{(a)} Stage-score distributions over 930 outcomes; thick and thin
  lines denote the interquartile and 10th--90th percentile ranges, and circles
  and diamonds denote medians and means. \textbf{(b)} Partial score versus the
  trimmed average turns in Table~\ref{tab:model-results}. \textbf{(c)} Mean
  tool-mediated time allocation across 200 sampled trajectories (50 per model),
  with phases coded from timestamped tool actions using fixed rules.}
  \label{fig:analysis-diagnostics}
\end{figure*}

\paragraph{Partial Progress Rarely Composes into Full Reproduction.}
Only 80 of 930 outcomes exceed the 90\% threshold, and these successes are
concentrated on 22 tasks; 71 tasks are never completed by any configuration,
and 11 of the remaining 22 are completed by at most two. Figure~
\ref{fig:analysis-diagnostics}(a) localizes this composition gap. Modeling and
Execution have median scores of 73.4\% and 73.5\%, respectively, whereas
Validation has a median of only 40.0\%. Even among the 304 outcomes scoring at
least 80\% on both Modeling and Execution, only 53 (17.4\%) exceed the
full-reproduction threshold, with a mean Validation score of 58.37\%. These
results show that
current agents can often reconstruct and execute substantial portions of a
scientific workflow, yet still fail to compose those partial successes into an
end-to-end reproduction that produces credible evidence and validates the
paper's claims.

\paragraph{Interaction Efficiency Varies Substantially Across Agents.}
Figure~\ref{fig:analysis-diagnostics}(b) compares each configuration's trimmed
mean interaction turns with its Partial score and finds only a weak negative
correlation ($r=-0.14$). Fable~5.1 and Fable~5 average only about 113 turns yet
achieve Partial scores of 58.17\% and 57.96\%, respectively, close to the
59.08\% attained by GPT-6-Astra with an average of 348.7 turns. By contrast,
DeepSeek-V4.1-Flash averages 372.6 turns but scores only 45.47\%. Longer
trajectories do not necessarily improve reproduction and may instead reflect
repeated trial and error, failure recovery, or continued execution of a
mis-specified workflow. These results indicate that interaction count alone is
insufficient to explain reproduction performance; the accuracy of early-stage
decisions may be an important factor shaping both interaction efficiency and
final scores.

\paragraph{Upstream Decisions Shape the Value of Execution.}
We sample 50 trajectories from each of four models and divide tool-mediated
time into paper reconstruction, implementation, native execution, diagnosis
and tuning, and evidence and replay. Figure~\ref{fig:analysis-diagnostics}(c)
shows that native execution accounts for 51.8--63.3\% of elapsed time, but
models differ substantially in how effectively they convert it into valid
scientific evidence. Fable~5 allocates 37.1\% of its time to paper
reconstruction and implementation, the largest upstream share among the four
models.

The F-stub UWB MIMO antenna task illustrates this pattern. Fable~5 scores
28.9\% in 159 turns, versus 27.4\% for GPT-6-Astra in 464 turns and 18.1\% for
HY3 in 390 turns. It completes three antenna variants and recovers evidence for
port reciprocity, radiation patterns, and diversity gain. This case shows that
execution alone is insufficient. Sound reconstruction, parameter choices, and
workflow design shape whether computation yields verifiable evidence.

\paragraph{Final Outputs Alone Cannot Distinguish Scientific Failure Modes.}
\label{sec:loss-audit}
Failures take two qualitatively different forms. Some workflows execute
successfully but support an incorrect conclusion because of errors in modes,
ports, parameters, mechanism implementation, or result extraction. On the
polarization-insensitive LNOI edge-coupler task, GPT-6-Astra receives full
Execution credit but only 41.9\% in Validation. Similarly, several models
complete replayable experiments for the eight-element smartphone MIMO antenna
but fail to reproduce its reported performance.

Other failures invalidate the evidence itself. Cached results can prevent
verifier-owned replay, privileged test-state information can invalidate
scientific results, and premature termination can leave no scoreable
submission. Final outputs obscure this distinction: one workflow executes but
reaches the wrong conclusion, whereas the other establishes no trustworthy
result.

Final-output agreement is therefore insufficient. By separately scoring
Modeling, Execution, and Validation and checking replay and provenance,
PAPERBENCHX distinguishes experimental-construction errors, incorrect
conclusions, and unverifiable evidence. End-to-end reproduction requires not
only an output, but one produced by a correctly constructed and independently
verifiable scientific workflow.

\flushbottom
\section{Conclusion}
\label{sec:conclusion}

\benchmarkname{} tests whether agents can turn published scientific claims into
executable workflows and independently verifiable evidence. Across 93 tasks in
12 research areas and 10 domain-native environments, we evaluate 10 frontier
agent configurations. The best achieves a $59.08\%$ partial score but only a
$13.98\%$ full rate; no configuration reaches full reproduction on 71 tasks.
Agents often build plausible models and run native simulations, but validation
remains their weakest stage, and more interaction does not necessarily improve
performance. These results make scientific reproduction an evidence-grounded
checkpoint for AI scientists and motivate better diagnosis, provenance, and
verification.

\clearpage
\section*{Contribution}

\paragraph{Core contributors.}
Ruoyu Wu\textsuperscript{1}, Zengji Tu\textsuperscript{1},
Aolong Sun\textsuperscript{1}, Tianyi Ma\textsuperscript{1},
Liang Chen\textsuperscript{1}, Kuan Li\textsuperscript{1}

\paragraph{Contributors.}
Junren Li\textsuperscript{1}, Xinbo Xu\textsuperscript{1}, Li Fu\textsuperscript{1},Ziyan Zhang\textsuperscript{1}, Pinrui Huang\textsuperscript{1}, Xiaochen Wang\textsuperscript{1}, Chengda Wu\textsuperscript{1}, Yutao Zhou\textsuperscript{1}, Jin Chen , Shen Yan , Xu Li , Zhongfei Hou , Nan Shu ,
Chaoyi Huang ,  Sihang Yuan ,
Zihan Xu

\paragraph{Advisors.}
Baobao Chang\textsuperscript{2}, Lin Chang\textsuperscript{2},
Jianjun Wu\textsuperscript{2}

\paragraph{Affiliations.}
\textsuperscript{1}UniPat AI \qquad
\textsuperscript{2}Peking University

\clearpage
\appendix
\setcounter{table}{0}
\renewcommand{\thetable}{A\arabic{table}}
\section{Task Package, Inventory, and Public Release}
\label{app:task-inventory}

\providecommand{\pbxgithubtask}[1]{\href{https://github.com/UniPat-AI/PaperBenchX/tree/7abb21c4ae83ddee7a05c4611e6428fa768ab6ce/tasks/#1}{\code{#1}}}

\subsection{Anatomy and Information Boundary of a Complete Task}
\label{app:task-anatomy}

We use the public \pbxgithubtask{06-gmr-grating-fano-meep} package to make the task
contract concrete. The package is part of the twelve-task release at
\href{https://github.com/UniPat-AI/PaperBenchX}{github.com/UniPat-AI/PaperBenchX}. The
task asks an agent to reproduce guided-mode resonances in a three-pair
multilayer dielectric grating using Meep. The starter submission and scientific
environment are agent-facing during a controlled trial, whereas the tests,
rubric, judge prompt, and reference artifacts remain in the evaluator
environment. All components are available for inspection.

\begin{lstlisting}[style=paperbenchxcode,basicstyle=\ttfamily\scriptsize]
gmr-grating-fano-meep/
|-- task.toml
|-- instruction.md
|-- environment/
|   |-- Dockerfile
|   |-- docker-compose.yaml
|   |-- init-workspace.sh
|   |-- starter_submission/
|   |   |-- README.md
|   |   |-- reproduce.sh
|   |   `-- src/
|   `-- paper/
|       |-- paper.md
|       |-- paper.pdf
|       |-- addendum.md
|       `-- paper_image/
`-- tests/
    |-- test.sh
    |-- evaluate.py
    |-- rubric.json
    |-- config/
    |   |-- judge_prompt.md
    |   `-- public_contract.md
    |-- refs/
    `-- src/
\end{lstlisting}

\paragraph{Components and evaluation-time visibility.}
Table~\ref{tab:task-package-components} separates the materials available to an
agent from evaluator-only resources. This boundary applies during evaluation;
the complete demonstration package, including its evaluator resources, is now
public for audit.

\begin{table}[htbp]
  \centering
  \small
  \setlength{\tabcolsep}{4pt}
  \renewcommand{\arraystretch}{1.08}
  \begin{tabularx}{\linewidth}{@{}p{0.24\linewidth}Xcc@{}}
    \toprule
    Component & Role & Agent & Evaluator \\
    \midrule
    Paper and addendum & Scientific scope, measurements, and evidence requirements & \cmark & \cmark \\
    Instruction & Goal, deliverables, and submission interface & \cmark & \cmark \\
    Environment & Simulator, dependencies, and resource envelope & \cmark & \cmark \\
    Starter submission & Editable source tree and replay entry point & \cmark & \cmark \\
    Rubric & Hierarchical, independently scored scientific requirements & \xmark & \cmark \\
    Judge prompt & Evidence-inspection and decision rules & \xmark & \cmark \\
    Reference artifacts & Expert-generated logs, arrays, and validation records & \xmark & \cmark \\
    \bottomrule
  \end{tabularx}
  \caption{Task components and their visibility during evaluation. Evaluator-only
  components are included in the public demonstration package after evaluation.}
  \label{tab:task-package-components}
\end{table}

\paragraph{Worked scientific contract.}
Table~\ref{tab:gmr-contract} summarizes the parts of the GMR task that determine
what must be executed and what evidence must be retained. The complete protocol
remains authoritative; this summary illustrates its structure rather than
replacing it.

\begin{table}[htbp]
  \centering
  \small
  \setlength{\tabcolsep}{4pt}
  \renewcommand{\arraystretch}{1.08}
  \begin{tabularx}{\linewidth}{@{}p{0.23\linewidth}X@{}}
    \toprule
    Field & Public contract \\
    \midrule
    Scientific target & Guided-mode resonances of a three-pair multilayer dielectric grating in Meep \\
    Cases & Unetched control plus full-depth air-slot widths of 0.030, 0.050, 0.070, and 0.150~$\mu$m \\
    Required evidence & Native spectra and logs, matched reference/device fluxes, complex reflected amplitudes, resonance fits, independent decay modes, and refinement records \\
    Replay interface & No-argument \code{bash /home/submission/reproduce.sh}, executed offline \\
    Compute envelope & Meep~1.17.1, 8 CPUs, 32~GiB RAM; 7 hours for the agent, 5 hours for native calculation, and 6 hours for replay and review \\
    Failure reporting & Missing, dark, overlapping, or non-identifiable features remain unresolved and cannot be replaced with paper values or invented poles \\
    \bottomrule
  \end{tabularx}
  \caption{Condensed public contract for the GMR demonstration task.}
  \label{tab:gmr-contract}
\end{table}

\paragraph{From contract to evidence leaves.}
The hidden rubric used during evaluation converts the public scientific
requirements into independent leaves. Representative leaves check whether
(i) the executed geometry contains the specified six layers and full-depth air
slots (Modeling), (ii) clean replay regenerates genuine Meep runs, raw arrays,
logs, and refinement controls (Execution), and (iii) power closure, pole fits,
mode agreement, and convergence satisfy the disclosed scientific conditions
(Validation). A missing item affects the leaves that depend on it rather than
silently imposing a global zero.

\paragraph{Evaluator policy.}
Only evidence regenerated by clean replay is admissible. The evaluator accepts
equivalent documented data layouts and unit conversions, but does not infer
missing units, phase, provenance, or scientific quantities from a report.
Scientific failures receive zero on the affected leaves; evaluator read or tool
failures instead withhold the judgment for repair and re-evaluation. Each leaf
is judged against its stated requirement without targeting a model-level score.

\paragraph{Complete public record.}
The pinned \pbxgithubtask{06-gmr-grating-fano-meep} directory contains the full
agent instruction, scientific addendum, environment definition, judge prompt,
rubric, and reference materials. 

\subsection{Complete Benchmark Inventory}
\label{app:complete-inventory}

Table~\ref{tab:complete-inventory} follows the 93-task result set used for the
reported model results. The numeric prefix identifies the
research area, and ``yes'' marks the public demo in that area.

\begingroup
\scriptsize
\setlength{\tabcolsep}{4pt}
\renewcommand{\arraystretch}{1.06}
\begin{longtable}{@{}p{0.09\linewidth}p{0.73\linewidth}p{0.10\linewidth}@{}}
\caption{Complete 93-task benchmark inventory used in the reported evaluation.}
\label{tab:complete-inventory}\\
\toprule
ID & Task & Public demo \\
\midrule
\endfirsthead
\multicolumn{3}{c}{\tablename~\thetable\ (continued)}\\
\toprule
ID & Task & Public demo \\
\midrule
\endhead
\midrule
\multicolumn{3}{r}{Continued on next page}\\
\endfoot
\bottomrule
\endlastfoot
\multicolumn{3}{@{}l}{\textit{01 RF antennas and arrays}}\\
1-1 & \code{5g-smartphone-eight-element-mimo} & \\
1-2 & \code{cpw-quad-port-uwb-mimo} & yes \\
1-3 & \code{dual-band-wlan-miniaturized-patch} & \\
1-4 & \code{f-stub-uwb-mimo} & \\
1-5 & \code{inkjet-flexible-dual-band-cp-patch} & \\
1-6 & \code{metasurface-broadband-cp-patch} & \\
1-7 & \code{parasitic-decoupling-two-monopole-mimo} & \\
1-8 & \code{tm30-high-gain-cp-patch} & \\
1-9 & \code{uwb-fractal-dgs-antenna} & \\
1-10 & \code{wideband-neutralization-line-mimo} & \\
\addlinespace
\multicolumn{3}{@{}l}{\textit{02 RF metasurfaces and scattering}}\\
2-1 & \code{anisotropic-coding-diffusion-metasurface} & yes \\
2-2 & \code{checkerboard-ebg-rcs-reduction} & \\
2-3 & \code{eight-circular-sector-wide-angle-absorber} & \\
2-4 & \code{fish-shaped-dual-function-polarization-converter} & \\
2-5 & \code{pcm-entropy-ga-coding-metasurface} & \\
2-6 & \code{uwb-pcm-checkerboard-rcs} & \\
\addlinespace
\multicolumn{3}{@{}l}{\textit{03 RF filters and frequency-selective surfaces}}\\
3-1 & \code{dual-passband-angular-stable-fss} & yes \\
3-2 & \code{dual-stopband-miniaturized-fss} & \\
3-3 & \code{embedded-window-rasorber} & \\
3-4 & \code{fourth-order-siw-bpf} & \\
3-5 & \code{quad-mode-wideband-microstrip-filter} & \\
\addlinespace
\multicolumn{3}{@{}l}{\textit{04 Integrated photonic devices}}\\
4-1 & \code{dualmode-tapered-coupler} & \\
4-2 & \code{eaton-lens-multimode-bend} & \\
4-3 & \code{heterogeneous-ln-sin-transitions-splitter} & \\
4-4 & \code{lnoi-multimode-bend} & \\
4-5 & \code{lnoi-pol-insensitive-edge-coupler} & \\
4-6 & \code{luneburg-pcw-coupler} & \\
4-7 & \code{mdm-sip-neff-mmi-design} & \\
4-8 & \code{mfe-multimode-crossing-soi} & \\
4-9 & \code{mmi-beamsplitter-soi} & \\
4-10 & \code{sin-bent-dc-splitter} & \\
4-11 & \code{soi-submicron-crossing} & \\
4-12 & \code{soi-swg-mmi-splitter} & \\
4-13 & \code{swg-anisotropic-phase-shifter} & yes \\
4-14 & \code{swg-mode-exchanger-yjunction} & \\
\addlinespace
\multicolumn{3}{@{}l}{\textit{05 Photonic crystals and topological photonics}}\\
5-1 & \code{half-w1-gainp-slow-light} & \\
5-2 & \code{hoti-quadrupole-corner} & \\
5-3 & \code{klein-tunneling-honeycomb-phc} & \\
5-4 & \code{ln-pti-bandstructure-edge-state} & \\
5-5 & \code{pc-line-defect-self-collimation} & \\
5-6 & \code{phc-fano-cavity-waveguide-gain} & \\
5-7 & \code{phc-superlens-imaging} & \\
5-8 & \code{phc-w1-slowlight-glide} & \\
5-9 & \code{pt-bragg-unidirectional-invisibility} & yes \\
5-10 & \code{self-collimation-hex-router} & \\
5-11 & \code{ssh-photonic-lattice-edge} & \\
5-12 & \code{topo-valley-ring-channel-filter} & \\
5-13 & \code{valley-bend-domainwall} & \\
5-14 & \code{zak-interface-1d-phc} & \\
\addlinespace
\multicolumn{3}{@{}l}{\textit{06 Optical metasurfaces, gratings, and plasmonics}}\\
6-1 & \code{al-wire-grid-polarizer-pitch-sweep} & \\
6-2 & \code{bowtie-aperture-resonance} & \\
6-3 & \code{fano-invisibility-dielectric-rod} & \\
6-4 & \code{gmr-grating-fano} & yes \\
6-5 & \code{integrated-metasurface-nanopillar-lookup} & \\
6-6 & \code{metagrating-anomalous-deflector} & \\
6-7 & \code{photonic-nanojet-twin-hook} & \\
6-8 & \code{plasmonic-dimer-hotspot} & \\
6-9 & \code{quasi-bic-metasurface-q} & \\
6-10 & \code{silicon-grating-antenna} & \\
\addlinespace
\multicolumn{3}{@{}l}{\textit{07 Optical computing and signal processing}}\\
7-1 & \code{brewster-spatial-differentiator} & yes \\
7-2 & \code{donn-i1-iris-classification} & \\
7-3 & \code{gmr-2d-edge-detector} & \\
7-4 & \code{optical-slab-spatial-integrator} & \\
7-5 & \code{phc-xor-logic-gate} & \\
7-6 & \code{swg-spatial-differentiator} & \\
\addlinespace
\multicolumn{3}{@{}l}{\textit{08 Optical resonators, filters, and tunable devices}}\\
8-1 & \code{clc-1d-resonator} & yes \\
8-2 & \code{lnoi-athermal-mwg-filter} & \\
8-3 & \code{sin-microring-optofluidic} & \\
8-4 & \code{soi-swg-pump-filter} & \\
8-5 & \code{vdw-mems-swx-switch} & \\
\addlinespace
\multicolumn{3}{@{}l}{\textit{09 Computational chemistry}}\\
9-1 & \code{ketene-water-chain} & \\
9-2 & \code{ni13c-ni2-kie-discrimination} & \\
9-3 & \code{nickel-catellani} & \\
9-4 & \code{nitrogen-deletion-dynamics} & \\
9-5 & \code{pimarenyl-bifurcation} & yes \\
9-6 & \code{ptrh-n2-activation} & \\
9-7 & \code{singh-sert-conformational-docking} & \\
9-8 & \code{solvent-bifurcation-dynamics} & \\
\addlinespace
\multicolumn{3}{@{}l}{\textit{10 Computational materials science}}\\
10-2 & \code{gong-be-elastic} & \\
10-3 & \code{haastrup-mos2-bands} & yes \\
10-5 & \code{choudhary-sic-properties-pbesol-hse06} & \\
10-6 & \code{tran-al-surface-energy} & \\
\addlinespace
\multicolumn{3}{@{}l}{\textit{11 Computational biology and bioinformatics}}\\
11-1 & \code{horlbeck-genetic-interactions} & yes \\
11-2 & \code{machina-metastatic-migration} & \\
11-3 & \code{moschandrea-mito-lipid-enterocyte} & \\
11-4 & \code{nichenet-ligand-target} & \\
11-5 & \code{ooka-wsme-l-folding} & \\
\addlinespace
\multicolumn{3}{@{}l}{\textit{12 Robotics and embodied AI}}\\
12-1 & \code{embodiedbench} & \\
12-2 & \code{mapgpt-r2r} & yes \\
12-3 & \code{octo} & \\
12-4 & \code{poliformer} & \\
12-6 & \code{rt1x} & \\
12-8 & \code{tracevla} & \\
\end{longtable}
\endgroup

\subsection{Public Demonstration Tasks}
\label{app:public-demos}

Table~\ref{tab:public-demos} lists the exact package identifiers and source
publications for the twelve packages in the public GitHub release. Raw
identifiers are retained because they map directly to the released task
directories at commit \code{7abb21c4}.

\begingroup
\scriptsize
\setlength{\tabcolsep}{3pt}
\renewcommand{\arraystretch}{1.08}
\begin{longtable}{@{}p{0.24\linewidth}p{0.27\linewidth}p{0.46\linewidth}@{}}
\caption{The twelve public demonstration tasks and their source publications.}
\label{tab:public-demos}\\
\toprule
Area & Task & Source publication and DOI \\
\midrule
\endfirsthead
\multicolumn{3}{c}{\tablename~\thetable\ (continued)}\\
\toprule
Area & Task & Source publication and DOI \\
\midrule
\endhead
\bottomrule
\endlastfoot
Radio-frequency antennas and arrays & \pbxgithubtask{01-cpw-quad-port-uwb-mimo} &
\textit{CPW Fed Compact UWB 4-Element MIMO Antenna with High Isolation};
\href{https://doi.org/10.3390/s21082688}{10.3390/s21082688} \\
Radio-frequency metasurfaces and scattering & \pbxgithubtask{02-anisotropic-coding-diffusion-metasurface} &
\textit{Broadband diffusion metasurface based on a single anisotropic element and optimized by the Simulated Annealing algorithm};
\href{https://doi.org/10.1038/srep23896}{10.1038/srep23896} \\
Radio-frequency filters and frequency-selective surfaces & \pbxgithubtask{03-dual-passband-angular-stable-fss} &
\textit{A Dual-Passband Frequency Selective Surface with High Angular Stability and Polarization Insensitivity};
\href{https://doi.org/10.3390/mi15060690}{10.3390/mi15060690} \\
Integrated photonic devices & \pbxgithubtask{04-swg-anisotropic-phase-shifter-meep} &
\textit{Ultra-broadband nanophotonic phase shifter based on subwavelength metamaterial waveguides};
\href{https://doi.org/10.1364/PRJ.373223}{10.1364/PRJ.373223} \\
Photonic crystals and topological photonics & \pbxgithubtask{05-pt-bragg-unidirectional-invisibility-meep} &
\textit{Unidirectional Invisibility induced by PT-Symmetric Periodic Structures};
\href{https://doi.org/10.1103/PhysRevLett.106.213901}{10.1103/PhysRevLett.106.213901} \\
Optical metasurfaces, gratings, and plasmonics & \pbxgithubtask{06-gmr-grating-fano-meep} &
\textit{Controlling Fano resonances in multilayer dielectric gratings towards optical bistable devices};
\href{https://doi.org/10.1038/s41598-018-34787-9}{10.1038/s41598-018-34787-9} \\
Optical computing and signal processing & \pbxgithubtask{07-brewster-spatial-differentiator-meep} &
\textit{Analog computing by Brewster effect};
\href{https://doi.org/10.1364/OL.41.003467}{10.1364/OL.41.003467} \\
Optical resonators, filters, and tunable devices & \pbxgithubtask{08-clc-1d-resonator-meep} &
\textit{Photonic eigenmodes and transmittance of finite-length 1D cholesteric liquid crystal resonators};
\href{https://doi.org/10.1038/s41598-023-43912-2}{10.1038/s41598-023-43912-2} \\
Computational chemistry & \pbxgithubtask{09-pimarenyl-bifurcation-v2} &
\textit{Post-transition State Bifurcations in Reactions That Form Pimarenyl Cation Diastereomers---Implications for Terpene Biosynthesis};
\href{https://doi.org/10.1021/jacs.5c16476}{10.1021/jacs.5c16476} \\
Computational materials science & \pbxgithubtask{10-haastrup-2018-mos2-bands} &
\textit{The Computational 2D Materials Database: high-throughput modeling and discovery of atomically thin crystals};
\href{https://doi.org/10.1088/2053-1583/aacfc1}{10.1088/2053-1583/aacfc1} \\
Computational biology and bioinformatics & \pbxgithubtask{11-horlbeck-2018-genetic-interactions} &
\textit{Mapping the Genetic Landscape of Human Cells};
\href{https://doi.org/10.1016/j.cell.2018.06.010}{10.1016/j.cell.2018.06.010} \\
Robotics and embodied AI & \pbxgithubtask{12-mapgpt-r2r} &
\textit{MapGPT: Map-Guided Prompting with Adaptive Path Planning for Vision-and-Language Navigation};
\href{https://doi.org/10.18653/v1/2024.acl-long.529}{10.18653/v1/2024.acl-long.529} \\
\end{longtable}
\endgroup

\subsection{Platforms and Resource Envelopes}
\label{app:demo-resources}

Table~\ref{tab:demo-resources} reports the limits declared by the public task
packages. The limits are configuration ceilings
rather than measured runtimes. An empty GPU entry means that no GPU count is
declared.

\begin{figure}[!htbp]
  \centering
  \includegraphics[width=\linewidth]{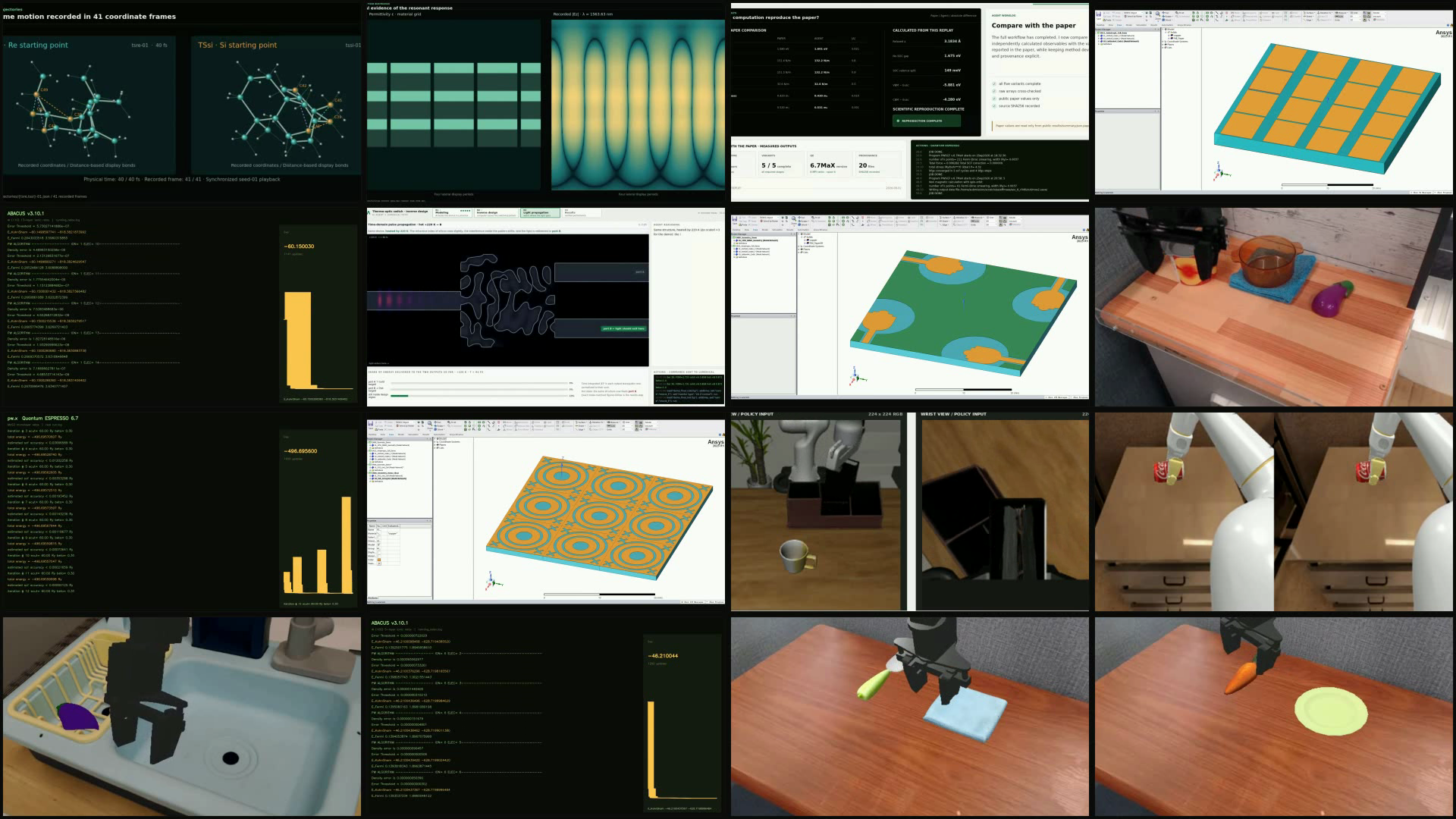}
  \caption{\textbf{Representative outputs from domain-native environments.}
  Panels show solver fields and spectra, numerical analysis outputs, and
  embodied-task scenes produced by scientific workflows in the benchmark.}
  \label{fig:simulation-environments}
\end{figure}

\begin{table}[htbp]
  \centering
  \scriptsize
  \setlength{\tabcolsep}{2.7pt}
  \renewcommand{\arraystretch}{1.08}
  \begin{tabularx}{\linewidth}{@{}lXrrrrrr@{}}
    \toprule
    ID & Execution stack & Agent h & Verifier h & CPU & RAM GiB & Disk GiB & GPU \\
    \midrule
    1-2 & Ansys HFSS & 24 & 25 & 32 & 128 & 200 & \\
    2-1 & Ansys HFSS & 72 & 73 & 32 & 128 & 200 & \\
    3-1 & Ansys HFSS & 48 & 49 & 32 & 128 & 300 & \\
    4-13 & Meep & 7 & 6 & 8 & 32 & 50 & \\
    5-9 & Meep & 5 & 5 & 8 & 32 & 50 & \\
    6-4 & Meep & 7 & 6 & 8 & 32 & 50 & \\
    7-1 & Meep & 6 & 5 & 8 & 32 & 50 & \\
    8-1 & Meep & 7 & 7 & 8 & 32 & 50 & \\
    9-5 & PySCF/GPU4PySCF & 7 & 7 & 8 & 32 & 50 & 1 \\
    10-3 & Quantum ESPRESSO & 7 & 7 & 8 & 32 & 50 & \\
    11-1 & Python scientific stack & 6 & 6 & 8 & 32 & 50 & \\
    12-2 & Local Qwen2.5-VL/vLLM & 18 & 7 & 8 & 32 & 100 & 1 \\
    \bottomrule
  \end{tabularx}
  \caption{Execution stacks and declared resource limits for the twelve public
  task packages in Table~\ref{tab:public-demos}}
  \label{tab:demo-resources}
\end{table}
\FloatBarrier

\section{Evaluation and Verification Protocol}
\label{app:benchmark}
\label{app:evaluation}
\label{app:protocol}

This appendix expands the evaluation protocol in
Section~\ref{sec:verification}. 
\subsection{Scope, Resources, and Trial Records}

\paragraph{Evaluation scope.}
The benchmark contains \nTasksCur{} tasks and \nLeaves{} scored rubric leaves.
All \nModels{} configurations are evaluated on the same \nResultTasks{} tasks,
with the fallback and missing-outcome policy defined in
Section~\ref{sec:experimental-setup}. The Partial score is the mean task score
over this fixed denominator. The Full rate counts tasks with an aggregate score
strictly greater than $0.9$; ties in reported rankings are resolved by
competition rank on unrounded scores.

\paragraph{Resource envelopes.}
Each task declares CPU, memory, storage, and wall-clock limits appropriate to
its scientific workflow. Across the benchmark, agent time limits range from
4 to 72 hours, with a median of 7 hours. Clean-replay and evaluator limits are
specified separately from the agent's development budget. Exact limits for the
twelve public demonstration tasks appear in Table~\ref{tab:demo-resources}.

\paragraph{Trial records.}
Each trial is linked to a task version, agent configuration, submitted package,
replay status, and per-leaf judgments. Agent runtime, clean-replay runtime, and
grading time are recorded separately. Missing telemetry remains missing rather
than being imputed, and infrastructure failures are distinguished from an
agent's failure to produce an executable submission.

\subsection{Expert Annotation and Release Review}

For each task, a domain expert maps the target claim to its source passage,
figure, or table; fixes the model configuration, experimental conditions,
observables, acceptable methodological alternatives, and required evidence;
and converts these requirements into independently scorable rubric leaves.
The public instruction, addendum, environment, and starter submission must
agree on the task scope and replay interface. Private reference workflows,
reference artifacts, and grading tolerances are withheld from agents, while
values reported in the source paper remain visible.

The private reference workflow is first executed to establish feasibility and
then rerun from a clean copy of the final environment through the same entry
point used for submissions. Calibration workflows representing successful,
partial, and invalid reproductions are scored to check that supported progress
receives credit and unsupported claims do not. Two independent domain experts
then review the scientific scope, fairness, reproducibility, evidence
requirements, and scoring behavior. Changes to the task contract, environment,
rubric, or evaluator require the affected downstream checks to be repeated
before release.

\subsection{Clean Replay and Evidence Grading}

\paragraph{Submission snapshot.}
The canonical entry point is
\code{bash /home/submission/reproduce.sh}. After delivery, the verifier records
the submitted source and required inputs, removes generated outputs and mutable
execution state, and reconstructs the workspace from the submitted package.
This prevents cached or pre-existing result files from being treated as evidence
from the replay.

\paragraph{Controlled execution.}
The package is executed with private evaluation resources inaccessible,
outbound network access disabled, and task-specific limits enforced. Exit
status, timing, source hashes, and available execution provenance are recorded.
These records establish how an artifact was produced; they do not by themselves
establish that its scientific content is correct.

\paragraph{Frozen evidence.}
Artifacts regenerated during replay are frozen as a read-only evidence set.
Deterministic checks cover directly testable properties such as required paths,
parseable formats, units, finite values, numerical consistency, sampling grids,
and task-specific tolerances. Simulator logs, raw arrays, trajectories,
convergence records, derived metrics, and visualizations are admitted only when
they can be associated with the current replay and the relevant task variant.

\paragraph{Leaf judgments and aggregation.}
Each rubric leaf is graded from its criterion, effective weight, relevant replay
artifacts, and deterministic check results. Requirements needing domain
interpretation are judged by the same GPT-5.6-sol evaluator used for every
reported configuration. Candidate reports may identify evidence but cannot
replace absent or contradictory artifacts. With leaf score $s_i$ and effective
weight $w_i$, the task score is $R=\sum_i w_i s_i$, where
$\sum_i w_i=1$. A missing artifact affects only the leaves that depend on it;
replay, infrastructure, and evaluator failures are recorded separately.

\subsection{Regression and Adversarial Checks}
\label{app:regression}

Task-specific regression tests implement the calibration step described in
Section~\ref{sec:tasks}. They pair valid controls with deliberately corrupted or
incomplete submissions to test whether the evaluator accepts supported evidence,
rejects unsupported claims, and confines a failure to the rubric leaves that
depend on it. Table~\ref{tab:threat-model} summarizes the principal check
families used across the released verifiers.

\begin{table*}[t]
  \centering
  \small
  \begin{tabular}{@{}p{0.25\textwidth}p{0.70\textwidth}@{}}
    \toprule
    Check family & Perturbation and expected behavior \\
    \midrule
    Contract and parser integrity & Preserve rubric hashes and complete leaf
    mappings; reject duplicate keys, nonfinite values, and invalid numerical
    grids while accepting documented equivalent formats. \\
    \addlinespace
    Native evidence and variant identity & Reject incomplete solver logs,
    mismatched setup names, explicit non-convergence, and copied variant outputs;
    accept a shared solve only when the task contract permits it and the variants
    remain identifiable. \\
    \addlinespace
    Evidence boundary & Exclude cached outputs and links outside permitted roots;
    keep private references and evaluator credentials inaccessible; fail closed
    when network isolation cannot be established. \\
    \addlinespace
    Completion and provenance & Give no completion credit to an empty submission;
    require delivery records, input hashes, and preprocessing conventions to agree
    with the task contract. \\
    \addlinespace
    Local failure accounting & Corrupt one input, metric, or figure panel and
    verify that only dependent rubric leaves lose credit. \\
    \addlinespace
    Interactive-task contract & Use evaluator-owned episode records and protocol
    identifiers; do not allow candidate summaries to replace required trajectory
    or visual evidence. \\
    \bottomrule
  \end{tabular}
  \caption{Regression cases used to check evidence admission and local scoring.
  Valid controls accompany invalid fixtures so that permissible variation is not
  mistaken for unsupported evidence.}
  \label{tab:threat-model}
\end{table*}

The twelve public task verifiers exercise different parts of this contract,
including numerical parsing, native solver evidence, provenance,
cached-output removal, and evaluator-owned records. Prompt-injection regression
cases are included to check whether instructions embedded in a candidate report,
log, or result field can alter the rubric or override a deterministic check.

\section{Additional Result Diagnostics}
\label{app:case-studies}

\subsection{Task Completion}

\begin{figure}[H]
  \centering
  \includegraphics[width=\textwidth]{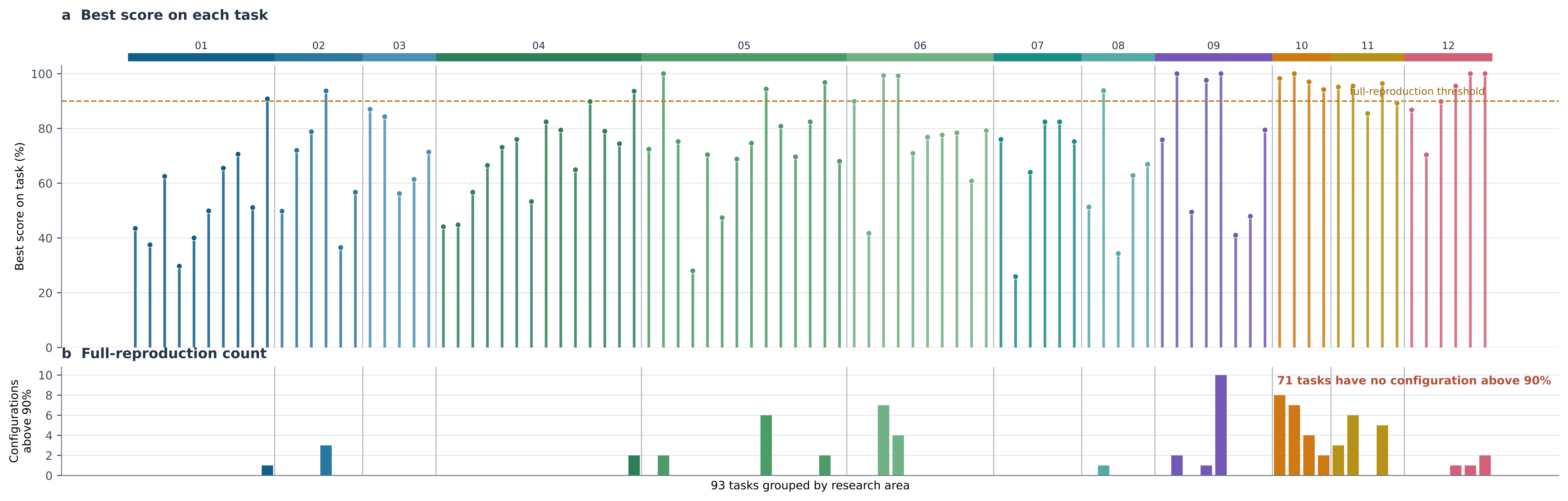}
  \caption{\textbf{Task-level completion landscape.}
  \textbf{(a)} Maximum Partial score attained on each of the 93 tasks, ordered
  by research area. \textbf{(b)} Number of configurations exceeding $90\%$ on
  the same task. The horizontal line marks the operational full-reproduction
  threshold; 71 tasks have no configuration above it.}
  \label{fig:task-completion-landscape}
\end{figure}

Figure~\ref{fig:task-completion-landscape} resolves the aggregate Full rate at
the task level. For each task, it reports the best score attained by any of the
ten configurations and the number of configurations exceeding the operational
$90\%$ threshold. The figure makes the concentration of complete reproductions
explicit: 71 tasks are not completed by any configuration.

\subsection{Cross-Simulator Trace Case Studies}

\begin{figure}[H]
  \centering
  \includegraphics[width=\textwidth]{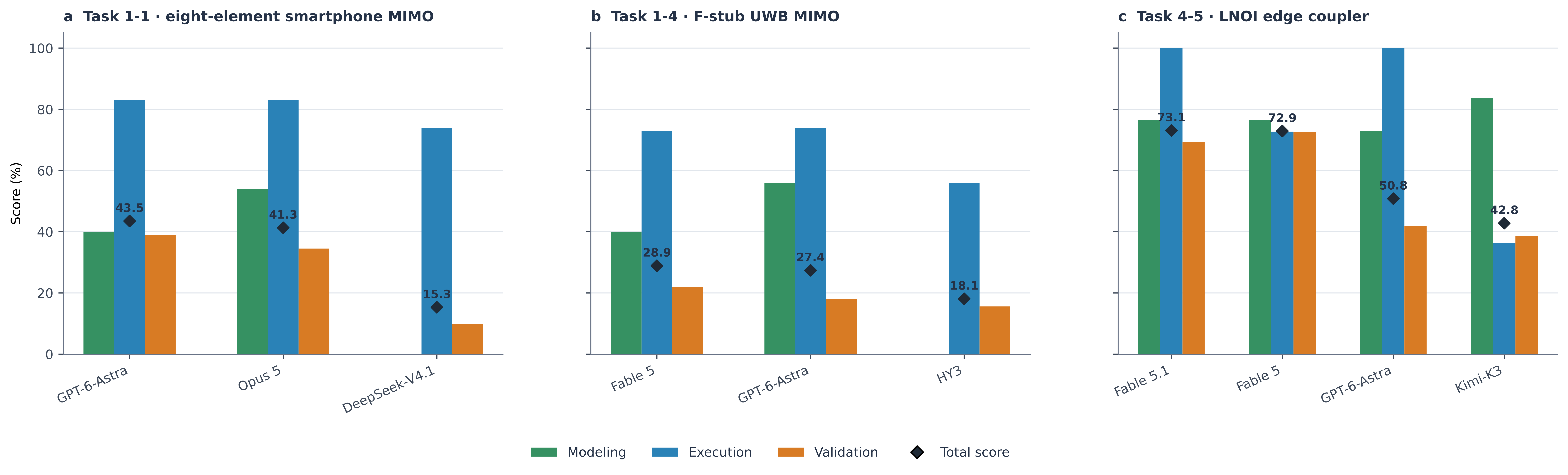}
  \caption{\textbf{Stage profiles in three trace case studies.}
  Bars report Modeling, Execution, and Validation scores; diamonds report the
  aggregate task score.}
  \label{fig:case-stage-profiles}
\end{figure}

This section expands the trace examples discussed in
Section~\ref{sec:loss-audit}. Figure~\ref{fig:case-stage-profiles} previews the stage-level
contrast developed in the three cases.

\paragraph{Eight-element smartphone MIMO (Task 1-1).}
The task requires two independent eight-port HFSS variants and evaluates
geometry, solver provenance and convergence, matching, isolation, efficiency,
ECC, gain, radiation patterns, and surface currents. GPT-6-Astra scores
$43.5\%$ in 187 turns; Opus~5 scores $41.3\%$ in 450 turns. Both replay native
HFSS workflows and recover several isolation,
ECC, gain, and current criteria. Opus~5 additionally records five and seven
adaptive passes, final maximum $\Delta S$ values of 0.0063 and 0.0078, and
independent meshes exceeding half a million tetrahedra.

Neither submission reproduces the central four-port matching target. Reflection
minima and the values at 3.45 GHz miss their required windows, and the baseline
and efficiency conditions fail. GPT-6-Astra also represents a critical length
as a reference parameter rather than using it in the model. Native execution
and agreement on secondary observables therefore do not establish the paper's
impedance and efficiency claims. DeepSeek-V4.1-Flash provides a complementary
efficiency example: 1,024 turns yield a score of $15.3\%$.

\paragraph{F-stub UWB MIMO (Task 1-4).}
The task evaluates plain, I-stub, and F-stub variants over a 501-point sweep.
As reported in the main text, Fable~5 scores $28.9\%$ in 159 turns, compared
with $27.4\%$ for GPT-6-Astra in 464 turns and $18.1\%$ for HY3 in 390 turns.
Fable~5 completes all three native variants and recovers evidence for port
reciprocity, radiation patterns, and diversity gain, but none of the three
submissions reproduces the principal S-parameter response. This comparison
supports the limited conclusion in Section~\ref{sec:discussion}: more
interactions do not necessarily improve reproduction when the model or
parameterization remains incorrect.

\paragraph{Polarization-insensitive LNOI edge coupler (Task 4-5).}
This workflow spans 39 Lumerical variants: FDE facet modes, TE/TM Gaussian
overlaps, geometry and alignment variants, and EME sweeps over four coupling
lengths. Fable~5.1 scores $73.1\%$ in 111 turns, followed by
Fable~5 at $72.9\%$ in 212 turns. Fable~5.1, Opus~5, and
GPT-6-Astra each receive full Execution credit, confirming that complete replay
is achievable in this environment.

The remaining errors are scientific. Fable~5.1 exceeds the allowed dual-tip TE
overlap, Fable~5 exceeds the facet-PDL limit, and GPT-6-Astra obtains an EME800
TM coupling of only 0.0474 despite complete replay artifacts. Kimi-K3 has the
highest Modeling score ($83.6\%$), but its EME output is NaN. The traces show why
intermediate checks matter: one facet solve, polarization and normalization,
and one EME case should be validated before the full sweep. NaNs, missing modes,
or incompatible matrix dimensions should stop downstream aggregation.

\paragraph{Synthesis.}
The cases explain why Modeling, Execution, and Validation are reported
separately; they are diagnostic examples rather than a cross-task ranking.

\end{document}